\documentclass[conference]{IEEEtran}
\usepackage{cite}
\usepackage{amsmath,amssymb,amsfonts}
\usepackage{algorithmic}
\usepackage{graphicx}
\usepackage{textcomp}
\usepackage{xcolor}
\def\BibTeX{{\rm B\kern-.05em{\sc i\kern-.025em b}\kern-.08em
    T\kern-.1667em\lower.7ex\hbox{E}\kern-.125emX}}

\usepackage{tikz}
\usepackage{tikzscale}
\usepackage{pgfplots}

\usepackage{booktabs}
\usepackage{mathtools}
\usepackage{amsfonts}

\newcommand{\set}[1]{\{{#1}\}}
\newcommand{\multiset}[1]{\{\!\{{#1}\}\!\}}
\newcommand{\WL}{\text{WL}}

\newcommand{\nats}{\mathbb{N}}

\newcommand{\colors}{\mathcal{C}}
\newcommand{\neighborhood}{\mathcal{N}}
\DeclarePairedDelimiter\size{\lvert}{\rvert}
\DeclareMathOperator{\hash}{hash}

\definecolor{nice_red}{rgb}{0.82, 0.1, 0.26}
\definecolor{nice_green}{rgb}{0.0, 0.5, 0.0}
\definecolor{nice_blue}{rgb}{0.0, 0.5, 1.0}

\begin{document}

\title{1-WL Expressiveness Is (Almost) All You Need}

\author{\IEEEauthorblockN{Markus Zopf}
		\IEEEauthorblockA{
				\textit{NEC Laboratories Europe GmbH} \\
				Heidelberg, Germany \\
				markus.zopf@neclab.eu}
	}


\maketitle

\begin{abstract}
It has been shown that a message passing neural networks (MPNNs), a popular family of neural networks for graph-structured data, are at most as expressive as the first-order Weisfeiler-Leman (1-WL) graph isomorphism test, which has motivated the development of more expressive architectures. In this work, we analyze if the limited expressiveness is actually a limiting factor for MPNNs and other WL-based models in standard graph datasets. Interestingly, we find that the expressiveness of WL is sufficient to identify almost all graphs in most datasets. Moreover, we find that the classification accuracy upper bounds are often close to 100\%. Furthermore, we find that simple WL-based neural networks and several MPNNs can be fitted to several datasets. In sum, we conclude that the performance of WL/MPNNs is not limited by their expressiveness in practice.
\end{abstract}

%

\section{Introduction}
Graph-structured data can be found in a wide range of domains such as chemistry, biology, physics, and computer science, where graphs are used, for instance, to model social networks, scientific citation graphs, protein structures, syntax trees, and the world wide web. Consequently, developing and understanding machine learning methods for graph-structured data has gained a lot of attention from the machine learning community \cite{Bronstein2021}. Many neural networks for graph-based datasets belong to the family of Message Passing Neural Networks (MPNNs) \cite{Scarselli2009}. Examples of popular MPNNs are the GraphSAGE model \cite{Hamilton2017}, Graph Convolutional Networks (GCNs) \cite{Kipf2017}, Graph Attention Networks (GATs) \cite{Velickovic2018}, and Graph Isomorphism Networks (GINs) \cite{Xu2019}.

\begin{figure}
	\centering
	\includegraphics[width=0.65\columnwidth]{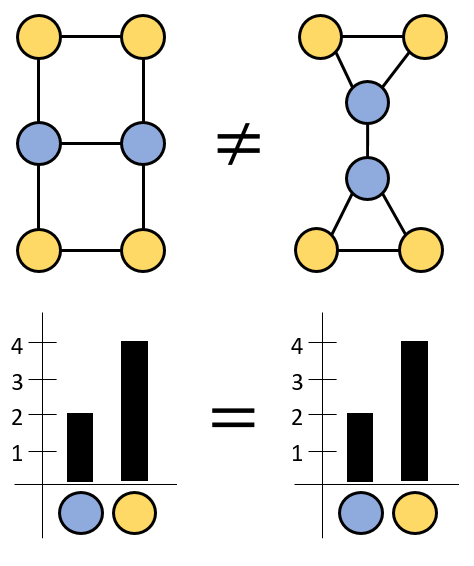}
	\caption{Example of two non-isomorphic graphs (top) that cannot be distinguished by the 1-WL graph isomorphism test since they have the same color/label histogram (bottom).}
	\label{fig:indistinguishable_graphs_example}
\end{figure}

Recently, it has been shown that the Weisfeiler-Leman (WL) graph isomorphism test \cite{Weisfeiler1968} provides an upper bound on the expressiveness of MPNNs \cite{Xu2019,Morris2019}, which means that MPNNs cannot distinguish graphs that cannot be distinguished by the WL graph isomorphism test even if they are not isomorphic. One example of two non-isomorphic graphs that cannot be distinguished by WL, and hence, MPNNs can be found in Figure~\ref{fig:indistinguishable_graphs_example}. Non-isomorphic indistinguishable graphs can be problematic for machine learning models since it is not possible for deterministic models to generate different predictions for indistinguishable graphs.

The limited expressiveness of WL and MPNNs has increased the interested in more expressive GNNs. For insistence, models that explicitly encode more expressive features into the graph representation \cite{Lee2019}, higher-order networks \cite{Lee2019,Maron2019,Morris2019}, approaches based on sub-graph sampling \cite{Bevilacqua2022,Bouritsas2020,Papp2021,Bodnar2021,Bodnar2021a}, and non-anonymous graphs \cite{Loukas2020,Loukas2020a,Abboud2021,Sato2021,Balcilar2021} have been developed.

However, a crucial question that remains open is whether the theoretically limited expressiveness of WL/MPNNs is actually a limiting factor for the generalization performance in real-world datasets \cite{Morris2020} or if synthetic counter-examples as illustrated in Figure~\ref{fig:indistinguishable_graphs_example} are rather rare cases and not relevant in practice. In this work, we address this question and find that the limited expressiveness is actually not a limiting factor in practice. More specifically, we find that a large fraction of graphs can be distinguished in most datasets by the WL test and that the theoretical upper bound performance is close to or even reaches a classification accuracy of 100\%. Moreover, we show that a simple WL-based machine learning model and several MPNNs can be fitted to the datasets, meaning that their limited expressiveness is not relevant in practice.

In sum, we conclude that WL and MPNNs are sufficiently expressive for many real-world datasets and that the development of new GNNs that are at most as expressive as the WL test should not be abandoned because of their theoretically limited expressiveness. Our findings align with the experimental results in \cite{Dwivedi2020}, where the authors find that MPNNs often outperform theoretically more expressive GNNs. It is noteworthy to mention that a sufficiently expressive model is no guarantee for a good generalization performance. Hence, it remains an open research questions which network architectures and training procedures will lead to a strong generalization performance for graph-structured data.

\begin{figure*}
	\centering
	\includegraphics[width=0.8\paperwidth]{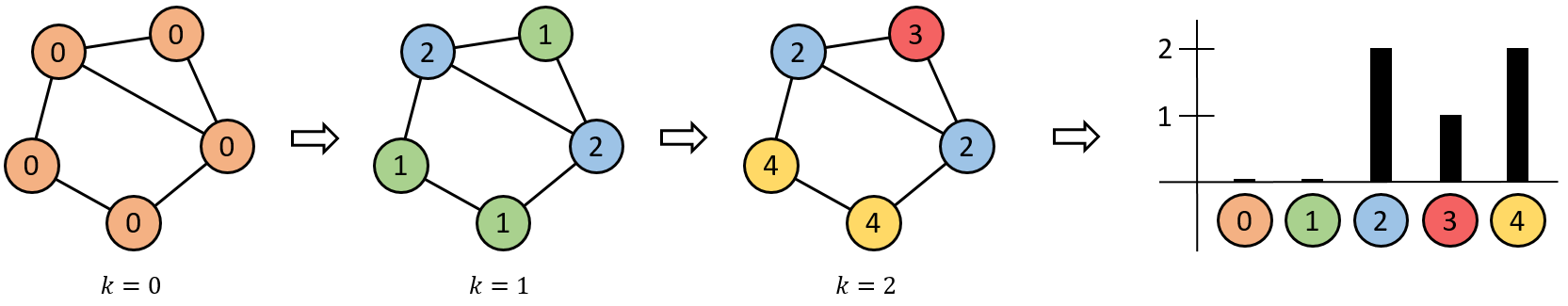}
	\caption{Illustration of the Weisfeiler-Leman graph color refinement algorithm. Initially ($k=0$), all nodes are labeled with the same color/label. In each iteration, the nodes are re-labeled according to an injective node labeling function based on the neighborhood nodes. For instance, the green and blue nodes obtained different colors in iteration $k=1$ since they have different neighborhoods (2 orange nodes vs. 3 orange nodes, respectively) in iteration $k=0$. After the maximum number of re-labeling iterations, the algorithm stops and a histogram based on the node color distributions in all iterations can be generated.}
	\label{fig:wl_example}
\end{figure*}

\section{Notation}
An \emph{undirected graph} is a pair $G = (V_G, E_G)$, where $V_G$ is a finite set of \emph{vertices} (also called nodes), and $E_G \subseteq \set{\set{u, v} : u, v \in V_G, u \neq v}$ is a symmetric, irreflexive, binary relation on $V_G$. The elements in $E_G$ are called \emph{edges}.

A \emph{colored undirected graph} is a pair $(G, \lambda)$ of an undirected graph $G$ and a \emph{coloring function} $\lambda: V_G \to \colors$, where $\colors$ is a set of colors (also called labels). Without loss of generality, we often use $\colors = \nats_0$ (i.e. the natural numbers including $0$). The coloring function $\lambda$ induces a partition $\pi(\lambda)$ of the graph nodes $V_G$. For a color $c \in \colors$, $\pi^{-1}(c) \subseteq V_G$ denotes the \emph{color class} of $c$, i.e. the set of nodes that are colored with $c$.

$\neighborhood(v) = \set{u : \set{u, v}, v \in V}$ denotes the \emph{neighborhood} of $v$, $\size{\cdot}$ denotes the size of a set, and $\multiset{\cdot}$ denotes a multiset, i.e. a set that may contain the same element multiple times.

\section{The Weisfeiler-Leman Graph Coloring Algorithm and Graph Isomorphism Test}
In the following, we first discuss the Weisfeiler-Leman graph coloring algorithm and then show how it can be used to implement a simple yet effective graph isomorphism test \cite{Weisfeiler1968}.

\subsection{The Weisfeiler-Leman graph coloring algorithm}
The first-order Weisfeiler-Leman graph coloring algorithm (1-WL, or WL for short) is an approach to generate a canonical representation of a graph as illustrated in Figure~\ref{fig:wl_example}. For an undirected graph $G$, the algorithm performs the following steps. First, all nodes in $G$ are assigned to the same initial color $c$. Without loss of generality, we use $0$ to indicate the first color and denote the initial coloring function by $\lambda^{(0)}(v)$ and set $\lambda^{(0)}(v) = 0, \forall v \in G_V$. For each iteration $k$, a coloring is defined according to

\begin{equation}
	 \lambda^{(k)}(v) = \hash(\lambda^{(k-1)}(v), \multiset{\lambda^{(k-1)}(u), u \in \neighborhood(v)}),
\end{equation}

where $\hash$ is a function that maps $v$'s previous color and the multiset of the previous colors of $v$'s neighbors to a color in $\colors$. As illustrated in Figure~\ref{fig:wl_example}, the coloring algorithm refines the node partition in each iteration (i.e. each color class can either be divided into multiple new color classes or stays constant). The coloring $\pi(\lambda^{(k)})$ is called \emph{stable coloring} if $\pi(\lambda^{(k)}) = \pi(\lambda^{(k+1)})$. In Figure~\ref{fig:wl_example}, a stable coloring has been generated for $k=2$. The algorithm terminates as soon as a stable coloring has been generated or a maximum number of $k$ iterations has been executed. While a naive implementation of the color refinement has quadratic running time of $O(m n)$, there are more efficient refinement strategies that run in $O(m \log n)$, where $n=\size{G_E}$ and $m=\size{G_V}$ \cite{Paige1987}. Instead of using $\lambda^{(0)}(v) = 0, \forall v \in G_V$ as initialization, predefined node labels can also be used as initialization if a dataset already provides labels for nodes (see Section~\ref{sec:analysis} for details).

Based on the obtained node colors, for each iteration $k$, a histogram of the node colors can be obtained according to

\begin{equation}
	h^{(k)}_G = \multiset{\lambda^{(k)}(v), v \in G_V}.
\end{equation}

The histogram $h^{(k)}_G$ can then be used as canonical graph representation. In practice, $h^{(k)}_G$ can be converted into a $d$-dimensional vector $v^{(k)}_G$ where $d$ equals the number of used colors. The entry in dimension $j$ is set according to 

\begin{equation}
	v^{(k)}_{G,j} = \size{\multiset{v : \lambda^{(k)}(v) = j, v \in G_V}}
\end{equation}

For a graph $G$, we denote the representation obtained by the first order WL coloring in iteration $k$ as 1-$\WL_G^{(k)}$, or $\WL^{(k)}$ for short. Consequently, the coloring in iteration $k=2$ for the graph in Figure~\ref{fig:wl_example} can be converted into the 5-dimensional vector $v^{(2)}_{G} = (0,0,2,1,2)^T$.

\subsection{The Weisfeiler-Leman graph isomorphism test}
The first-order Weisfeiler-Leman graph isomorphism test uses the previously defined WL-based canonical graph representation to test if two graphs are isomorphic. Two undirected graphs $G = (V_G, E_G)$ and $H = (V_H, E_H)$ are said to be \emph{isomorphic} if there is a bijection $\varphi: E_G \to E_H$ such that

\begin{equation}
	\set{v, u} \in E_G \iff \set{\varphi(v), \varphi(u)} \in E_H \forall v, u \in V_G
\end{equation}
Two colored undirected graphs $(G, \lambda_G)$ and $(H, \lambda_H)$ are \emph{isomorphic} if $G$ and $H$ are isomorphic and $\lambda_G(v) = \lambda_H(\varphi(v)) \forall v \in V_H$.

The Weisfeiler-Leman graph isomorphism test iteratively computes graph representations for two graphs $G$ and $H$ and iteration $k \in \set{1, \dots, K}$ according to the WL graph coloring algorithm. If the graph representations $h^{(k)}_G$ and $h^{(k)}_K$ differ in iteration $k$, the algorithm terminates. In this case, it is guaranteed that $G$ and $H$ are not isomorphic. The algorithm also terminates if a stable coloring has been generated for both graphs (i.e. the coloring does not change anymore) or the number of iteration reaches a predefined maximum number of iterations $k$. In these cases, it is not guaranteed that the graphs are not isomorphic, since the WL algorithm is known to not be able to distinguish all non-isomorphic graphs. In this case, the graphs are said to be \emph{indistinguishable} by the WL test and we write $G \approx H$ to express that $G$ and $H$ cannot be distinguished by the WL test. Given a set of graphs $\mathfrak{G}$, a graph $G \in \mathfrak{G}$ is said to be \emph{identifiable by WL} if there is no other graph $H \in \mathfrak{G}, H \neq G$ that is not distinguishable from $G$ by the WL test.

\section{Expressiveness of WL and MPNNs}
Several machine learning method that are based on WL-representations have already been proposed \cite{Morris2021}. When WL-representations are used as a basis for machine learning models, the fact WL cannot distinguish all non-isomorphic graphs can become problematic since it can lower the best achievable accuracy of the model.

In general, the expressiveness of a machine learning model (also called model capacity) is a crucial property that describes how well a model can fit a given dataset, for instance expressed in terms of the Vapnik–Chervonenkis (VC) dimension \cite{Vapnik1971}. If a model is not sufficiently expressive for a specific dataset, it is impossible to properly fit it to the dataset. A prominent example of insufficient expressiveness are linear classification models that are trained on data that requires a non-linear decision boundary. 

The expressiveness of a model needs to be distinguished from the generalization abilities of a model, which describes the ability of a model to generalize to unseen data instances and is usually estimated by applying a trained model on an unseen test dataset. While a sufficient expressiveness is necessary to achieve a good generalization performance (i.e. a good performance on an unseen test set), it is not sufficient to achieve a good generalization performance. Hence, a strong expressiveness does not guarantee a good generalization performance. In fact, it is easy in many cases to create sufficiently expressive models that do not generalize well. 

As already mentioned, the WL test is not able to distinguish all non-isomorphic graphs. In the context of WL-based graph representations, this means that non-isomorphic graphs may be represented with the same color histogram / vector representation. The graphs illustrated in Figure~\ref{fig:indistinguishable_graphs_example} are an example for this situation. In general, it has been shown that WL cannot distinguish non-isomorphic graphs is several scenarios \cite{Kiefer2020,Sato2021a}.

However, there are also some positive results about the expressiveness of WL. For instance, it has been shown that WL can distinguish all non-isomorphic trees \cite{Kiefer2020}. Moreover, it has been shown that WL is powerful enough to distinguish almost all pairs of randomly generated non-isomorphic graphs except for rare counterexamples. This means that the fraction of graphs that cannot be identified in a set of randomly generated graphs with $n$ nodes tends to 0 when $n \to \infty$ \cite{Babai1980}. Even stronger expressiveness has been shown for higher-dimensional WL algorithms. For instance, it has been shown that the 2-WL test identifies almost all $d$-regular graphs for every degree $d$ \cite{Bollobas1982}.

The limited expressiveness of the WL test and the graph representations generated by the WL graph coloring algorithm are not only relevant for graph isomorphism testing but also when the WL graph representations are used in machine learning models \cite{Morris2021}, for instance in kernel machines \cite{Shervashidze2009}, since their generalization performance can be limited by the expressiveness of the WL representations. More specifically, (deterministic) machine learning models that use WL-based representations to represent graphs for graph classification cannot generate different predictions for graphs that cannot be distinguished by the WL test. More formally, for two (potentially non-isomorphic) graphs $G$ and $H$ with $G \approx H$ (i.e. $h_G = h_H$), it is not possible to learn a function $f$ such that $f(h_G) \neq f(h_H)$, which is problematic when non-isomorphic graphs have the same 1-WL representation but a different class label. For instance, if graphs with a triangular pattern are labeled positive and graphs without a triangular pattern are labeled negative, no WL-based machine learning function can obtain a better prediction accuracy than 0.5 for the two graphs in Figure~\ref{fig:indistinguishable_graphs_example}, since both graphs are either labeled positive or negative, but it is impossible to label one graph positive and the other graph negative. Hence, we obtain an upper bound of 0.5 for the prediction accuracy.

Recently, it has been shown that the limited expressiveness of 1-WL also limits the expressiveness of a popular family of graph neural networks (GNNs). More specifically, it has been shown that GNNs that are based on the message passing approach (i.e. MPNNs \cite{Scarselli2009}) are at most as expressive as the 1-WL test \cite{Xu2019,Morris2019}. Key to this insight is the fact that the message passing algorithm can be interpreted as a continuous version of the 1-WL graph coloring algorithm. Among the MPNNs with limited expressiveness are popular GNNs such as  GraphSAGE, GCN, GIN, and GAT. Among these networks, the GIN architecture is specifically interesting since it has been shown to be exactly as expressive as 1-WL, while other MPNNs are potentially even less expressive \cite{Xu2019}. Moreover, \cite{Garg2020} showed that several graph properties such as shortest/longest cycle, diameter, and certain motifs cannot be computed by GNNs that rely entirely on local information. They also provide the first data dependent generalization bounds for MPNNs. \cite{Chen2020} studies limitations of MPNNs to count graph substructures. \cite{Chen2019} draw a connection between graph isomorphism testing and the approximation of permutation-invariant functions.

\section{Towards More Expressive Architectures}
The limited expressiveness of the WL isomorphism test and GNNs based on the message passing algorithm has motivated the development of new approaches for machine learning on graph-structured data that are more expressive, i.e. that do not suffer from the limited expressiveness of the WL test. In the following, we describe several works and categorize them into four different groups, namely methods that use more expressive features, GNNs that use higher-order structures, graph sub-sampling based approaches, and non-anonymous networks.

The first set of approaches to extend the expressiveness of MPNNs beyond WL expressiveness  explicitly encode more expressive features into the graph representation. For instance, a motif \cite{Milo2002} detector can be used to analyze input graphs and to add information about the presence or absence of specific motifs or the number of specific motifs to the graph representation. A simple way to achieve this is to combine the canonical representation obtained by the WL coloring algorithm and the additionally obtained motif features. An example of such a combination of different features can be found in \cite{Ben-Hur2005}. The obtained representation is more expressive than the initial WL representation and can, for instance, distinguish the two graphs in Figure~\ref{fig:indistinguishable_graphs_example}. \cite{Lee2019} propose an extension of this idea by learning a weighted sum of motif features up to a specific size.

The second way to improve the 1-WL test is to use higher-order structures such as pairs of nodes, or even larger groups. The resulting architectures are called higher-order networks. \cite{Maron2019} show that k-order graph neural networks are as expressive as the k-WL test. For instance, a 3-order GNNs is as expressive as the 3-WL test and, hence, more expressive than 1-WL \cite{Huang2021}. However, higher-order GNNs come with a higher computational cost since they process higher order tensors and the number of different higher-order structures can grow quickly when larger structures are considered. Moreover, \cite{Maron2019} present an extended 2-order GNN (which is computationally easier to handle) that is as expressive as 3-WL. \cite{Morris2019} propose $k$-GNNs which are based on $k$-WL and take higher-order structures into account.

Another direction to obtain more expressive GNNs is to sample sub-graph structures, i.e. by removing edges and/or nodes from the original input graph. For instance, Equivariant Subgraph Aggregation Networks (ESAN) \cite{Bevilacqua2022} improve the expressiveness of MPNNs by modeling each graph as a set of subgraphs. \cite{Bouritsas2020} propose Graph Substructure Networks (GSNs) that use additional substructure-based features during the message passing. DropGNNs \cite{Papp2021} can also be grouped into the sub-graph structure methods since they sample sub-graphs by randomly dropping some nodes. Message Passing Simplicial Networks (MPSNs) perform message passing over simplicial complexes (SCs), a specific class of sub-graph structures. CW Networks \cite{Bodnar2021} extend MPSNs \cite{Bodnar2021a} as they use the broader class of closure-finite weak (CW) complexes.

Another direction to extend the expressiveness of GNNs is to convert graphs into non-anonymous networks, i.e. graphs in which each node has an attribute that uniquely identifies it. Examples of such attributes are unique node ids or (unique) random features. \cite{Loukas2020} shows that GNNs are Turing universal under sufficient conditions on their architecture (depth and width), node attributes (i.e. (partial) non-anonymity), and layer expressiveness, which means that they can solve the graph isomorphism problem. \cite{Loukas2020a} introduced the communication capacity measure that measures how much information the nodes of a network can exchange during the forward pass. Moreover, it introduces communication complexity, which addresses the question how much information needs to be exchanged, e.g. to distinguish (connected) graphs and trees. Moreover, it has been shown that GNNs with random node initialization (RNI) are universal \cite{Abboud2021}, meaning that they can approximate any function on graphs of any fixed order. Consequently, GNNs with RNI are more expressive than WL \cite{Sato2021}.


\section{Analysis of WL-Expressiveness in Real-World Datasets}
\label{sec:analysis}
\begin{table}
	\centering
	\caption{Overview of popular graph datasets}
	\label{tab:dataset_details}
	\begin{tabular}{l r c c c}
		Dataset	& Size & Node labels & Classes & Unique\\
		\toprule
		DD			& 1,178		& 89 	& 2		&  1.00 \\ 
		ENZYMES		& 600		& 3 	& 6		&  0.98 \\ 
		MUTAG		& 188		& 7 	& 2		&  0.87 \\
		NCI1		& 4,110		& 37 	& 2		&  0.97 \\
		NCI109		& 4,127		& 38 	& 2		&  0.97 \\
		PROTEINS	& 1,113		& 3 	& 2		&  0.94 \\ 
		\midrule      
		COLLAB		& 5,000		& 0 	& 3		&  0.78 \\
		IMDB-B		& 1,000		& 0 	& 2		&  0.42 \\
		IMDB-M		& 1,500		& 0 	& 3		&  0.19 \\
		REDDIT-B	& 2,000		& 0 	& 2		&  1.00 \\
		REDDIT-M-5K	& 4,999		& 0 	& 5		&  1.00 \\
	\end{tabular}
\end{table}

In the following, we analyze if and to which extend the limited expressiveness of the first-order Weisfeiler-Leman graph isomorphism test (i.e. 1-WL), and hence, the limited expressiveness of message passing neural networks is a limiting factor in popular graph datasets that have been used by prior works that proposed more expressive GNNs \cite{Papp2021,Bodnar2021a}. To this end, we compute the fraction of graphs that can be identified by the 1-WL test with different number of iterations $k$ (i.e. 1-$WL^{(k)}$). Moreover, we compute an upper bound for the classification accuracy that can be achieved with machine learning models that are based on 1-WL representations and message passing neural networks.

Table~\ref{tab:dataset_details} provides an overview of the datasets that are used in this study. We report the size of the datasets in terms of number of graphs in the dataset. Moreover, we report the number of different node labels in case that the nodes in the graph have been initially labeled. Initial node labels are relevant in the study since they can be used as initialization when generating graph representations with WL. We also report how large the fraction of unique graphs is, i.e. for how many graphs no other isomorphic graph is present in the dataset. The fraction of unique graphs is important since it is a dataset-specific limitation for the fraction of identifiable graphs. Interestingly, we found that only 19\% and 42\% of the graphs in IMDB-M and IMDB-N are unique, respectively.

\subsection{Fraction of WL-indentifiable graphs}
In the following, we evaluate how severe the limited expressiveness of the 1-WL representations, and thus also how severe the limited expressiveness of MPNNs, is. To this end, we compute the fraction of non-isomorphic graphs in several commonly used datasets that can be identified by WL representations. More specifically, we compute for each graph WL-based representations 1-$\WL^{(k)}$, where $k$ indicates the number of iterations of the WL algorithm. A small fraction of identifiable graphs means that the WL algorithm cannot distinguish many graphs. In the extreme case, a fraction of $0\%$ means that all graphs have the same 1-WL encoding. Vice versa, a large fraction of identifiable graphs shows that only a few graphs cannot be distinguished. A fraction of $100\%$ means that each unique graph has an individual representation.

\begin{table}
	\centering
	\caption{Fraction of identifiable graphs for $k$ WL iterations}
	\label{tab:individual_wl_representations}
	\begin{tabular}{l c c c}
		Dataset	& \multicolumn{3}{c}{Identifiable Graphs} 	\\
					& k=1		&	k=2		&	k=3		\\
		\toprule
		DD			&	100.00	&	100.00	&	100.00	\\
		ENZYMES		&	100.00	&	100.00	&	100.00	\\
		MUTAG		&	32.32	&	92.68	&	96.34	\\
		NCI1		&	94.18	&	99.47	&	100.00	\\
		NCI109		&	94.91	&	99.40	&	100.00	\\
		PROTEINS	&	100.00	&	100.00	&	100.00	\\
		\midrule     
		COLLAB		&	100.00	&	100.00	&	100.00	\\
		IMDB-B		&	100.00	&	100.00	&	100.00	\\
		IMDB-M		&	100.00	&	100.00	&	100.00	\\
		REDDIT-B	&	100.00	&	100.00	&	100.00	\\
		REDDIT-M-5K	&	100.00	&	100.00	&	100.00	\\
	\end{tabular}
\end{table}

We report the results of the experiment in Table~\ref{tab:individual_wl_representations} and observe that a large fraction of graphs are identifiable in many datasets. More specifically, more than 94\% of the graphs can be identified in 10 out of 11 datasets with just a single iteration of WL. In 8 out of 11 datasets, already all non-isomorphic graphs can be identified with one iteration of the WL graph coloring algorithm. Only MUTAG is an outlier with only $32.32\%$ of identifiable non-isomorphic graphs. More iterations (i.e. $k > 1$) increase the fraction of identifiable graphs in 3 dataset (MUTAG, NCI1, and NCI109). Interestingly, the number of identifiable graphs is already at its maximum in all datasets that do not have node labels. One explanation for this observation is the fact that the variety of different graph representations is smaller when all nodes are initialized with the same color in contrast to graphs that are initialized with a large set of different node labels/colors. In sum, we conclude that the limited expressiveness of WL is not a crucial limitation in most of the datasets. Hence, the expressiveness of WL is usually not a limitation for the generalization performance.

\subsection{Upper performance bounds for WL-based representations}
In the previous section, we computed the fraction of WL-identifiable graphs, which provides insights on the expressiveness of WL without considering the class labels of the graphs. However, in practice, it is more relevant to estimate an upper bound or the classification accuracy, which indicates the best possible performance a deterministic model can achieve. As already discussed above, no deterministic\footnote{Models with a non-deterministic behavior may achieve higher performance. However, GNNs are usually implemented as deterministic functions.} MPNN with the expressiveness of the corresponding WL representation can achieve a better performance than a WL-based model. The classification upper performance bound is often higher than the fraction of identifiable instances. The reason for this is the fact that graphs that have the same label do not necessarily have to be identified by a machine learning model in order to correctly classify the graphs. It can even be desired that graphs that belong to the same class are represented in the same way to obtain a higher generalization performance. 

In the following, we estimate an upper bound for the classification accuracy of machine learning models that are based on WL representations. To this end, we compute the WL representation for all graphs for $k$ iterations. For each cluster of graphs with the same WL representation, we perform a majority voting to obtain the best possible prediction for this cluster. As already discussed above, the same class needs to be predicted for all graphs in a specific cluster.

\begin{table}
	\centering
	\caption{Upper bound accuracy for $k$-layer WL representation.}
	\label{tab:correctly_classified}
	\begin{tabular}{l c c c}
		Dataset	& \multicolumn{3}{c}{Upper bound performance} 	\\
		& k=1		&	k=2		&	k=3		\\
		\toprule
		DD			&	100.00	&	100.00	&	100.00	\\
		ENZYMES		&	100.00	&	100.00	&	100.00	\\
		MUTAG		&	95.74	&	99.47	&	100.00	\\
		NCI1		&	99.54	&	99.81	&	99.83	\\
		NCI109		&	99.73	&	99.85	&	99.88	\\
		PROTEINS	&	99.73	&	99.73	&	99.73	\\
		\midrule     
		COLLAB		&	97.98	&	97.98	&	97.98	\\
		IMDB-B		&	88.60	&	88.60	&	88.60	\\
		IMDB-M		&	63.27	&	63.27	&	63.27	\\
		REDDIT-B	&	100.00	&	100.00	&	100.00	\\
		REDDIT-M-5K	&	100.00	&	100.00	&	100.00	\\
		
	\end{tabular}
\end{table}

Table~\ref{tab:correctly_classified} reports the results of the experiment. We observe that a very high accuracy can in theory be achieved by WL-based models even for $k=1$.  In 4 datasets (DD, ENZYMES, REDDIT-B, and REDDIT-M-5K), the upper performance bound for $k=1$ is even 100\%, which means that a WL-based model with $k=1$ is sufficiently expressive to perfectly classify the dataset. In the MUTAG dataset, the difference between the fraction of identifiable graphs and the upper performance bound is most prominent. Even though only a small fraction of 32.32\% of the graphs are WL identifiable, a WL-based model can still achieve an accuracy of 95.74\%. Important to note is that the highest achievable performance is lower than 100\% in datasets with non-unique graphs such as IMDB and COLLAB \cite{Ivanov2019}.

\subsection{Results without using initial node labels}
As mentioned before, some datasets come with predefined node features. However, it is also common to evaluate the performance of GNNs if the node features are not used. In this case, all node features that are present in the dataset are replaced with the same constant label. Hence, it is also interesting to evaluate if WL representations are sufficiently expressive in this case. To this end, we perform the same analysis as before, but replace the provided node labels with the same constant label such that all nodes are labeled identically.

\pgfplotstableread[row sep=\\,col sep=&]{
	dataset & wLabels & woLabels \\
	DD      & 100.00  & 100.0 \\
	ENZ     & 100.00  & 97.12 \\
	MUT     & 32.32   & 16.51 \\
	NCI1    & 94.18   & 40.31 \\
	NCI109  & 94.91   & 40.82 \\
	PROT    & 100.00  & 88.36 \\
}\dataone

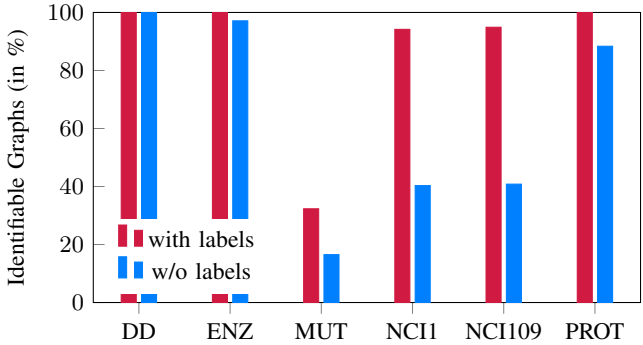
\begin{figure}
	\begin{tikzpicture}
		\begin{axis}[
			ybar,
			bar width=.2cm,
			width=\columnwidth,
			height=.3\textwidth,
			legend pos=south west,
			legend style={draw=none},
			symbolic x coords={DD,ENZ,MUT,NCI1,NCI109,PROT},
			xlabel near ticks,
			ylabel near ticks,
			ymin=0,ymax=100,
			xtick pos=left,
			ytick pos=left,
			style = {font=\small},
			ylabel={Identifiable Graphs (in \%)},
			]
			\addplot[nice_red,fill=nice_red] table[x=dataset,y=wLabels]{\dataone};
			\addplot[nice_blue,fill=nice_blue] table[x=dataset,y=woLabels]{\dataone};
			\legend{with labels, w/o labels}
		\end{axis}
	\end{tikzpicture}
	\caption{Fraction of identifiable graphs with and w/o node features for $k=1$.}
	\label{fig:wo_node_features_k1}
\end{figure}

\pgfplotstableread[row sep=\\,col sep=&]{
	dataset & wLabels & woLabels \\
	DD      & 100.00  & 100.0 \\
	ENZ     & 100.00  & 100.0 \\
	MUT     & 92.68   & 75.23 \\
	NCI1    & 99.47   & 97.85 \\
	NCI109  & 99.40   & 97.34 \\
	PROT    & 100.00  & 100.0 \\
}\datatwo

\begin{figure}
	\begin{tikzpicture}
		\begin{axis}[
			ybar,
			bar width=.2cm,
			width=\columnwidth,
			height=.3\textwidth,
			legend pos=south west,
			legend style={draw=none},
			symbolic x coords={DD,ENZ,MUT,NCI1,NCI109,PROT},
			xlabel near ticks,
			ylabel near ticks,
			ymin=0,ymax=100,
			xtick pos=left,
			ytick pos=left,
			style = {font=\small},
			ylabel={Identifiable Graphs (in \%)},
			]
			\addplot[nice_red,fill=nice_red] table[x=dataset,y=wLabels]{\datatwo};
			\addplot[nice_blue,fill=nice_blue] table[x=dataset,y=woLabels]{\datatwo};
			\legend{with labels, w/o labels}
		\end{axis}
	\end{tikzpicture}
	\caption{Fraction of identifiable graphs with and w/o node features for $k=2$.}
	\label{fig:wo_node_features_k2}
\end{figure}
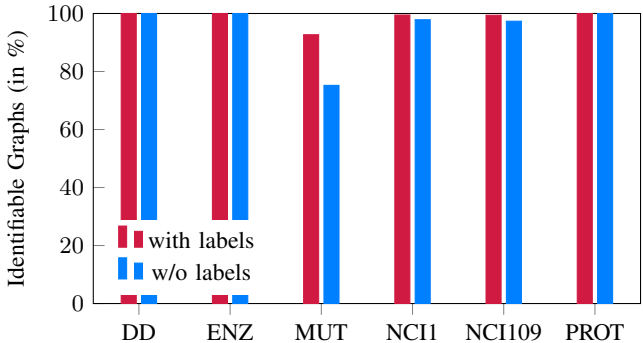

A comparison of the fraction of identifiable graphs with and without node features for $k=1$ and $k=2$ can be found in Figure~\ref{fig:wo_node_features_k1} and Figure~\ref{fig:wo_node_features_k2}, respectively. For $k=1$, we can see that replacing the original node labels reduces the fraction of identifiable graphs in several datasets. However, the situation already changes when the number of iterations is increased to $k=2$. In this case, we see that the differences between using and not using node labels decreases. There are also datasets such as DD and ENZYMES where the node labels do not contribute to the identifiability of the graphs.

Moreover, we report the upper bound for the classification accuracy without using node features in Table~\ref{tab:individual_wl_representations_wo_node_features}. Even though the number of identifiable graphs decreases substantially when no node features are used for $k=1$, the upper bound performance is still very high, which means that a very high classification accuracy can be achieved even without using node features. For $k=2$, the difference becomes even smaller.

\begin{table}
	\centering
	\caption{WL upper bound performance without using node features}
	\label{tab:individual_wl_representations_wo_node_features}
	\begin{tabular}{l c c c}
		Dataset	& \multicolumn{3}{c}{Upper Bound Performance} 	\\
		& k=1		&	k=2		&	k=3		\\
		\toprule
		DD			&	100.00	&	100.00	&	100.00	\\
		ENZYMES		&	98.83	&	100.00	&	100.00	\\
		MUTAG		&	91.49	&	96.28	&	96.81	\\
		NCI1		&	85.21	&	99.22	&	99.42	\\
		NCI109		&	85.51	&	98.91	&	99.32	\\
		PROTEINS	&	95.24	&	97.48	&	97.48	\\
	\end{tabular}
\end{table}

\subsection{WL expressiveness and upper accuracy bounds for $k=0$}
Prior works have shown that a good generalization performance can be achieved for graph classification even if the structure of the graph is not considered \cite{Errica2020}. A simple classification model that does not use the graph structure can be obtained when all node features are aggregated without performing any message passing operations. This idea corresponds to the WL algorithm for $k=0$ since no re-coloring is performed for $k=0$.

\begin{table}
	\centering
	\caption{Identifiable graphs and upper bound performance for $k=0$}
	\label{tab:individual_wl_representations_and_performance_k0}
	\begin{tabular}{l c cc}
		Dataset & Node Labels & Identifiable &	Upper Bound Acc.	\\
		\toprule
		DD			& yes	&	99.32	&	100.00 \\
		ENZYMES		& yes	&	43.39	&	81.33 \\
		MUTAG		& yes	&	25.00	&	93.09 \\
		NCI1		& yes	&	56.34	&	91.34 \\
		NCI109		& yes	&	56.98	&	91.66 \\
		PROTEINS	& yes	&	52.86	&	91.91 \\
		\midrule    
		DD			& no	&	15.11	&	83.70 \\
		ENZYMES		& no	&	3.05	&	38.50 \\
		MUTAG		& no	&	0.92	&	86.17 \\
		NCI1		& no	&	0.28	&	63.70 \\
		NCI109		& no	&	0.33	&	63.46 \\
		PROTEINS	& no	&	5.51	&	73.23 \\
		\midrule                          		
		COLLAB		& no	&	2.22	&	60.70 \\
		IMDB-B		& no	&	3.80	&	60.60 \\
		IMDB-M		& no	&	3.82	&	44.13 \\
		REDDIT-B	& no	&	19.39	&	83.85 \\
		REDDIT-M-5K	& no	&	9.06	&	55.17 \\
	\end{tabular}
\end{table}

We report the fraction of identifiable graphs and the upper bound performance with and without using node features for $k=0$ in Table~\ref{tab:individual_wl_representations_and_performance_k0}. We can see that the fraction of identifiable graphs decreases substantially for $k=0$ when no node features are available, which is reasonable since the WL representation reduces to a representation based on the number of nodes in the graphs. The effect is less severe when node features are available, but still clearly observable when compared with the results for $k=1$. Hence, we conclude that using initial node features and at least one iteration of the WL algorithm have a substantial positive effect on the WL identifiability. However, the effect on the upper bound classification performance is less strong even for $k=0$, specifically when node features are used. DD is an extreme case in which the upper performance bound does not drop below 100\%. In ENZYMES, the effect is stronger and the upper bound drops from 100\% to 81.33\%. In sum, we conclude that upper performance bound is still reasonably high for $k=0$ when using node features.

\subsection{WL-MLP and GNN upper performance bounds}
So far, we have focused on the WL expressiveness and the upper bound performance which answers the question if we need more expressive GNNs for standard graph datasets from a theoretical point of view. In the following, we also evaluate how well popular GNNs can fit the graph datasets, which not only considers the expressiveness of the GNNs but also their trainability. Trainability of a machine learning model describes how easy it is to fit a model to a specific dataset. Since the models are trained with gradient-based approaches that do not guarantee to find the global optimum in the parameter space, it is unlikely that we obtain the optimal upper bound performance but only an approximation thereof. Similarly, we only obtain an approximation of the expressiveness.

\begin{table}
	\centering
	\caption{Accuracy upper bound for WL-MLP and different GNNs}
	\label{tab:wl_gnn_trained_expressiveness}
	\begin{tabular}{l c c c c}
		Dataset 	& WL-MLP 	& GCN 		& GAT 		& GIN	\\
		\toprule
		DD			& 100.00&	100.00	&	100.00	&	100.00 \\
		ENZYMES		& 100.00&	90.67	&	84.83	&	99.50 \\
		MUTAG		& 95.74	&	90.96	&	91.49	&	100.00 \\
		NCI1		& 99.54	&	97.27	&	96.23	&	99.34 \\
		NCI109		& 99.69	&	76.86	&	78.63	&	88.08 \\
		PROTEINS	& 99.73	&	76.55	&	76.91	&	92.63 \\
		\midrule	                        	        
		COLLAB		& 97.92	&	71.70	&	52.00	&	76.18 \\
		IMDB-B		& 88.60	&	66.00	&	51.00	&	73.70 \\
		IMDB-M		& 63.20	&	37.93	&	34.33	&	59.53 \\
		REDDIT-B	& 100.00&	92.45	&	51.00	&	93.45 \\
		REDDIT-M-5K	& 59.67	&	50.59	&	21.06	&	57.33 \\
	\end{tabular}
\end{table}

We report the empirical upper bound performance of the models in Table~\ref{tab:wl_gnn_trained_expressiveness}. In this experiments, we used only one iteration of WL and single layer GNNs. Hence, the results can be compared with the results in Table~\ref{tab:correctly_classified}, column $k=1$. First, we find that the WL-based multi-layer perpetration baseline (WL-MLP) can successfully learn the datasets with node features, meaning that WL is sufficiently expressive for these datasets. Interestingly, the results are very close to the previously computed upper bound results in Table~\ref{tab:correctly_classified}, meaning that the WL-MLP is not only very expressive, but can also be fitted easily. The results for the GNNs are not as high as WL-MLP, but they still demonstrate a strong expressiveness. In the datasets without node features, it is more difficult to observe a high expressiveness of the machine learning models in many cases. This aligns with the results in Table~\ref{tab:correctly_classified}, in which we found that the limited expressiveness is most severe in the IMDB-B and IMDB-M datasets. The performance in REDDIT-M-5K is surprisingly low for all models. As mentioned above, the obtained results are approximations of the upper bound performance and better results might be achieved with an improved training procedure.

\subsection{Motif histograms}
Motif-based representations are a popular alternative to WL-based representation since they can capture properties of graphs that cannot be captured by WL representations. In the following, we evaluate motif-based representations to see if they are a reasonable choice in standard graph datasets. However, due to the high computational cost to count motifs, counting them is not feasible in several datasets. Hence, we only provide results for a subset of the datasets in Table~\ref{tab:individual_motif_representations_and_performance}.

\begin{table}
	\centering
	\caption{Identifiable graphs and upper bound performance for motifs up to size 4}
	\label{tab:individual_motif_representations_and_performance}
	\begin{tabular}{l c c}
		Dataset & Identifiable &	Upper Bound Acc.\\
		\toprule
		ENZYMES		& 99.49	& 99.83 \\
		MUTAG		& 15.85	& 92.55 \\
		NCI1		& 17.12	& 75.52 \\
		NCI109		& 17.08	& 75.09 \\
		\midrule
		IMDB-B		& 100.00	& 88.60 \\
		IMDB-M		& 100.00	& 63.27 \\
	\end{tabular}
\end{table}

The results show that the fraction of unique identifiable graphs and the upper bound performance for motif-based representations is lower than WL representations with $k=1$ for the datasets with node features. For datasets without node features, we don't observe substantial differences between WL representations and motif-based representations. Similarly, we observe that the upper performance bounds for datasets with node features is lower than for WL-based representations with $k=1$ in NCI1 and NCI109. However, the performance is only slightly lower in ENZYMES and MUTAG. Since motif-based representation are usually shorter, presuming a better generalization performance could be reasonable.

\section{Conclusions}
A main motivation for the development of several new machine learning methods for graph-structured data is the limited expressiveness of the WL graph isomorphism test and MPNNs. However, our experiments show that WL-based representations are sufficiently expressive to distinguish most graphs in several datasets. Already one iteration of WL was sufficient to obtain a large fraction of WL-identifiable graphs. Moreover, we found that WL is sufficiently expressiveness to achieve a near perfect classification accuracy in almost all datasets. Moreover, a simple WL-based neural network and popular GNNs such as the GIN can be fitted to most datasets. Hence, we conclude that the limited expressiveness of MPNNs is not a limiting factor for their generalization performance and more expressive GNNs are not necessary to achieve a good performances. Hence, our findings suggest that the development of new MPNNs whose expressiveness is limited by the WL is still relevant and should not be abandoned. Moreover, observing a higher generalization performance of more expressive GNNs cannot be attributed to their stronger expressiveness, but can likely be explained by other factors.

Furthermore, we found that two 1-WL iterations can lead to a larger fraction of identifiable graphs than only one iteration in a few cases, but the added value of a third iteration is marginal. Furthermore, we found that only one 1-WL iteration is sufficient to achieve a high upper bound of the classification accuracy. Both results indicate that one or two iterations of WL are already sufficiently expressive in practice. We also found that motif-based representations are less strong than one iteration of WL and also have a lower upper bound accuracy, which is a disadvantage of motif-based representations in addition to their much higher computational costs.

\bibliographystyle{IEEEtran}
\bibliography{references}

\end{document}